\documentclass[conference]{IEEEtran}
\IEEEoverridecommandlockouts

\usepackage{cite}
\usepackage{comment}
\usepackage{amsmath,amssymb,amsfonts}
\usepackage{algorithmic}
\usepackage{graphicx}
\usepackage{textcomp}
\usepackage{xcolor}
\def\BibTeX{{\rm B\kern-.05em{\sc i\kern-.025em b}\kern-.08em
    T\kern-.1667em\lower.7ex\hbox{E}\kern-.125emX}}
\begin{document}

\title{A System for Train Condition Monitoring and Structural Health Assessment of Rail Vehicles}

\author{
\IEEEauthorblockN{Maximilian Posner}
\IEEEauthorblockA{\textit{Institute of Machine Components} \\
\textit{University of Stuttgart}\\
Stuttgart, Germany \\
maximilian.posner@ima.uni-stuttgart.de}
\and
\IEEEauthorblockN{Martin Dazer}   
\IEEEauthorblockA{\textit{Institute of Machine Components} \\
\textit{University of Stuttgart}\\
Stuttgart, Germany \\
martin.dazer@ima.uni-stuttgart.de}
\and
\IEEEauthorblockN{Daniela Lauer}   
\IEEEauthorblockA{\textit{AutomatedTrain} \\
\textit{DB InfraGO AG}\\
Berlin, Germany \\
daniela.lauer@deutschebahn.com}
\and
\IEEEauthorblockN{Robert Winkler-Höhn}
\IEEEauthorblockA{\textit{Institute of Vehicle Concepts} \\
\textit{German Aerospace Center}\\
Stuttgart, Germany \\
robert.winkler-hoehn@dlr.de}
\and
\IEEEauthorblockN{Mathilde Laporte}
\IEEEauthorblockA{\textit{Institute of Vehicle Concepts} \\
\textit{German Aerospace Center}\\
Stuttgart, Germany \\
Mathilde.Laporte@dlr.de}
\and
\IEEEauthorblockN{Tobias Herrmann}
\IEEEauthorblockA{\textit{Niederlassung Berlin} \\
\textit{Institut für Bahntechnik GmbH}\\
Berlin, Germany \\
he@bahntechnik.de}
\and
\IEEEauthorblockN{Martin Köppel}   
\IEEEauthorblockA{\textit{AutomatedTrain} \\
\textit{DB InfraGO AG}\\
Berlin, Germany \\
martin.koeppel@deutschebahn.com}
}

\maketitle

\begin{abstract}

The ongoing digitalization of rail systems and the increasing use of artificial intelligence (AI) are fundamentally transforming the design, operation, and maintenance of rail vehicles. While fully automated operation at Grade of Automation 4 (GoA4) is well established in metro systems, its deployment in mainline rail remains limited. This is primarily due to stringent safety requirements and the complexity of open operational environments. Current perception systems based on cameras, radar, and lidar are effective in detecting objects but provide limited capability for reliably identifying impacts, collisions, and driving-over events.
This paper presents a novel approach for real-time vehicle condition monitoring and impact detection that integrates structural sensor technologies with AI-based data analysis. The proposed framework addresses three key applications: (1) automated detection of impacts, structural damage, and driving-over events, (2) condition-based maintenance enabled by continuous monitoring, and (3) long-term data analytics to support vehicle design optimization. The results demonstrate the feasibility of the proposed approach and highlight its potential to enhance operational safety, enable predictive maintenance strategies, and support the transition toward fully automated operation in mainline rail systems
\end{abstract}


\section{Introduction}

The digital transformation of the railway sector and the integration of Artificial Intelligence (AI) are driving profound changes in the design, operation, and maintenance of rail vehicles. Advanced sensor technologies, combined with AI-based data analysis, open up new possibilities for product development, life-cycle optimization, and the realization of automated operations. A crucial aspect is the integration of dedicated sensor systems that enable continuous condition monitoring, event detection, and intelligent decision-making.

Current rail vehicles are only sparsely equipped with such systems, which limits their ability to support higher levels of automation. To enable safe and efficient operation, especially in the context of fully automated train control, vehicles must be capable of detecting impacts, structural damage, collisions, and driving over events, while simultaneously providing reliable information for condition-based maintenance and long-term fleet optimization.

The KI-MeZIS (German acronym for: AI-Methods for Condition Monitoring and Need-driven maintenance for Rail Vehicle) project addressed these challenges by combining AI methods with novel vehicle-side sensors. Three cases were investigated: (1) detection of impacts, damage, and driving-over events, (2) development of condition-based maintenance strategies based on continuous data collection, and (3) long-term data analysis for optimized vehicle design. This paper presents the results for all of the three use cases. 

\begin{figure}
    \centering
    \includegraphics[width=1\linewidth]{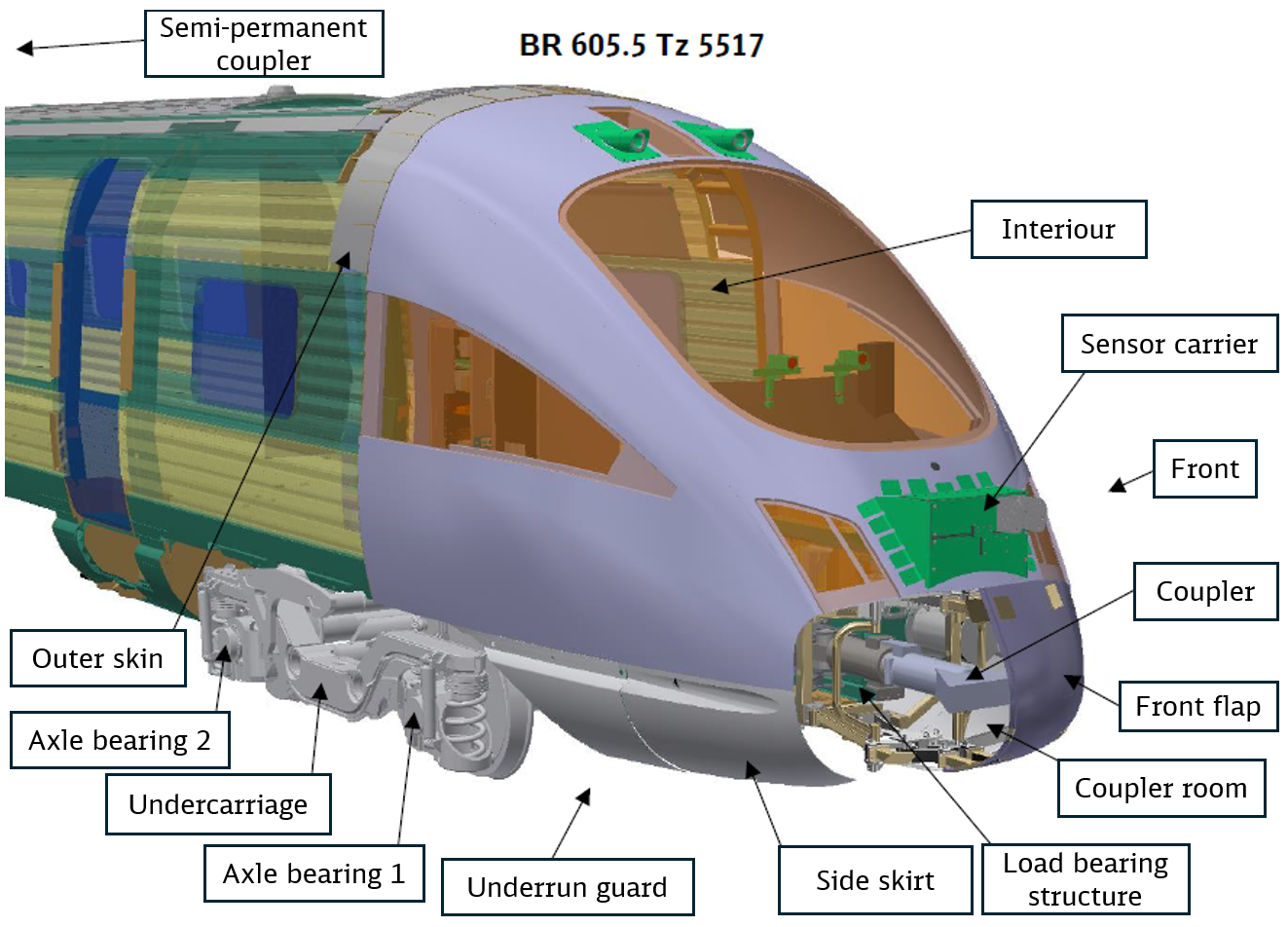}
    \caption{Equipped train structures on the aTL.}
    \label{fig:aTL_Train}
\end{figure}

\section{State-of-the-Art}
Digitization, lightweight design, and advanced sensing technologies are driving innovation in the railway sector. Within the concern program "Digitale Schiene Deutschland" \cite{DSD}, sensor-based event detection and monitoring were investigated as essential for operational safety and reliability. The program provides specifications and functional architectures for the future rail system. Laboratory and field tests validate the feasibility of various sensor concepts. For that, test trains such as the advanced Train Lab (aTL) \cite{DB_aTL} can be utilized.

Until now, fully automated rail operations have been predominantly implemented in closed metro systems. In contrast, automation in mainline railways remains limited, mainly due to more stringent safety requirements and the challenges associated with open operational environments. In such environments, external objects may enter the running path and therefore must be reliably detected and considered in train operation. In Germany, the EBA research report published in 2019 \cite{EBA} provides a comprehensive overview of projects related to digital driver assistance, digitization, and automation. At the international level, the UITP World Report on Metro Automation \cite{WorldReportOnMetroAutomation} documents the increasing global adoption of driverless metro systems. Beyond metro applications, a prominent example is Rio Tinto’s GoA4 freight trains in Australia, which operate fully autonomously on the regular rail network \cite{RioTinto_1, RioTinto_2}.

Lightweight design approaches, such as those investigated in the faWaSiS project \cite{WinklerHoehn2023} and related studies \cite{Dittus2013, Koenig2013}, demonstrate the potential of fiber-reinforced composite structures for highly loaded rail vehicle components. These materials offer high strength at reduced weight. The integration of Structural Health Monitoring (SHM) systems uses embedded or surface-mounted sensors. Such systems enable early damage detection and support predictive maintenance. They also facilitate digital twin applications. At the same time, fire safety and long-term durability requirements are fulfilled.

To test technologies, the aTL, a re-equipped high-speed train for real-world testing, serves as a testing platform \cite{DB_aTL}.

\section{Installed Sensors on the Train}
Within the KI-MeZIS project, several different sensors were installed in the aTL.
The installation of the system was carried out in close cooperation between DB, DB InfraGO AG, DB Systemtechnik and MSG Ammendorf. 

The sensors were selected and placed to support both structural monitoring and collision detection. Hence, piezoelectric acceleration, strain gauge sensors (DMS), air pressure, and localization sensors were identified as suitable for capturing reliable high-quality data in real operations that can be utilized to fulfill the use cases considered.

Figure \ref{fig:aTL_Train} gives a schematic overview of the equipped train structures.
Uniaxial and triaxial (Althen FRAB-10, FLAB-10) strain gauges sensors were installed on the underrun guard, the undercarriage, the load bearing structures, and the semi-permanent coupler of the aTL to monitor local deformations. They enable early detection of structural damage and, despite requiring protective encapsulation, offer a cost-efficient solution with high sensitivity and detailed material insights.

Several uniaxial and triaxial piezoelectrical acceleration sensors (Kistler 8715B1K0S00T, 8766A250AB, 8705A250M1) were added to capture impacts on the vehicle as well as driving over events and were installed on the front flap, the side skirt, the coupler, the underrun guard, the axle bearing and the undercarriage. These sensores have a measuring range of ±250g or ±1000g. They were mounted with adhesive adapters (Kistler), avoiding drilling or structural modifications.

For the strain and acceleration sensors, the sensor positions were determined using finite element (FE) simulations. Strain gauges were placed in high-strain and stress-concentration areas to capture the probable deformation during collisions. Acceleration sensors were installed in structurally representative, protected and accessible locations.

Air pressure sensors (Endevco Model 8510B-500-60-D) were mounted on the train’s outer skin and on the front sensor carrier of the train, quantifying aerodynamic loads. These measurements enable the calculation and analysis of the aerodynamic forces on the vehicle. 
Rotary encoders (Knorr Bremse, Deuta DF16) are installed on the front and rear axles of the vehicle. These measure the rotational speed of the wheels.
Furthermore, a GNSS system installed on the train was utilized to determine the geographic coordinates of the train for event evaluation, for example tunnels, rail switch, stations etc.
Temperature sensors (RS Pro Typ K131-4751, RS Pro Typ K726-118) were installed inside the train and in the coupler room of the train, to measure the influence of the weather on the sensor signals. Finally, a RGB camera (Net C-GP4206M-100-4) was installed in the front sensor carrier to observe the track area. The camera data served as ground truth data in case an event was visible in the signal data but not annotated during the test ride.

In total, the following sensors have been installed: 10 uniaxial and 2 triaxial acceleration sensors, 3 uniaxial and 9 triaxial strain sensors, 2 rotary sensors, 14 air pressure sensors, 2 temperature sensors and 1 camera. 
Furthermore, a 3D printed cable channel was installed on the train to transport the cables from the trains front area to the interior of the train.
To record the signals two coupled data loggers (DEWE3-M4) from DEWETRON were utilized.
The installed system on the train was approved according to the EU fourth railway package 
\cite{Europarl2016}.

\section{Data Aquisition}

Sensor data was collected during experiments, simulations and test drives in order to systematically record for both regular operation and non-regular events. For all laboratory experiments a similar sensor setups as on the train has been used, so all results were interoperable and could be used for training machine learning (ML) algorithms. 

\subsection{Test Drives}

The aTL, was used in the regular railway network to record data under normal operating conditions. No uncommon events were expected during this process. This data were used to record the characteristic signals of regular journeys and to create reference signals for different types of routes.
A total of 22 measurement days were performed on the regular network during live operation. During these test runs, care was taken to ensure that a wide range of test scenarios were taken into account:
(1) High-speed section were used. The train can travel at maximum speed (55 m/s) on these sections.
(2) Tunnel trips were carried out, since the vehicle is subjected to high air pressure when entering a tunnel.
(3) Secondary railway lines were used. The quality of the rails on these routes is usually poorer than on the main tracks because these lines are serviced and maintained less frequently.
(4) Different weather conditions were captured to see if the results are weather resistant.

\subsection{Driving-over Tests}

\begin{figure}
    \centering
    \includegraphics[width=1\linewidth]{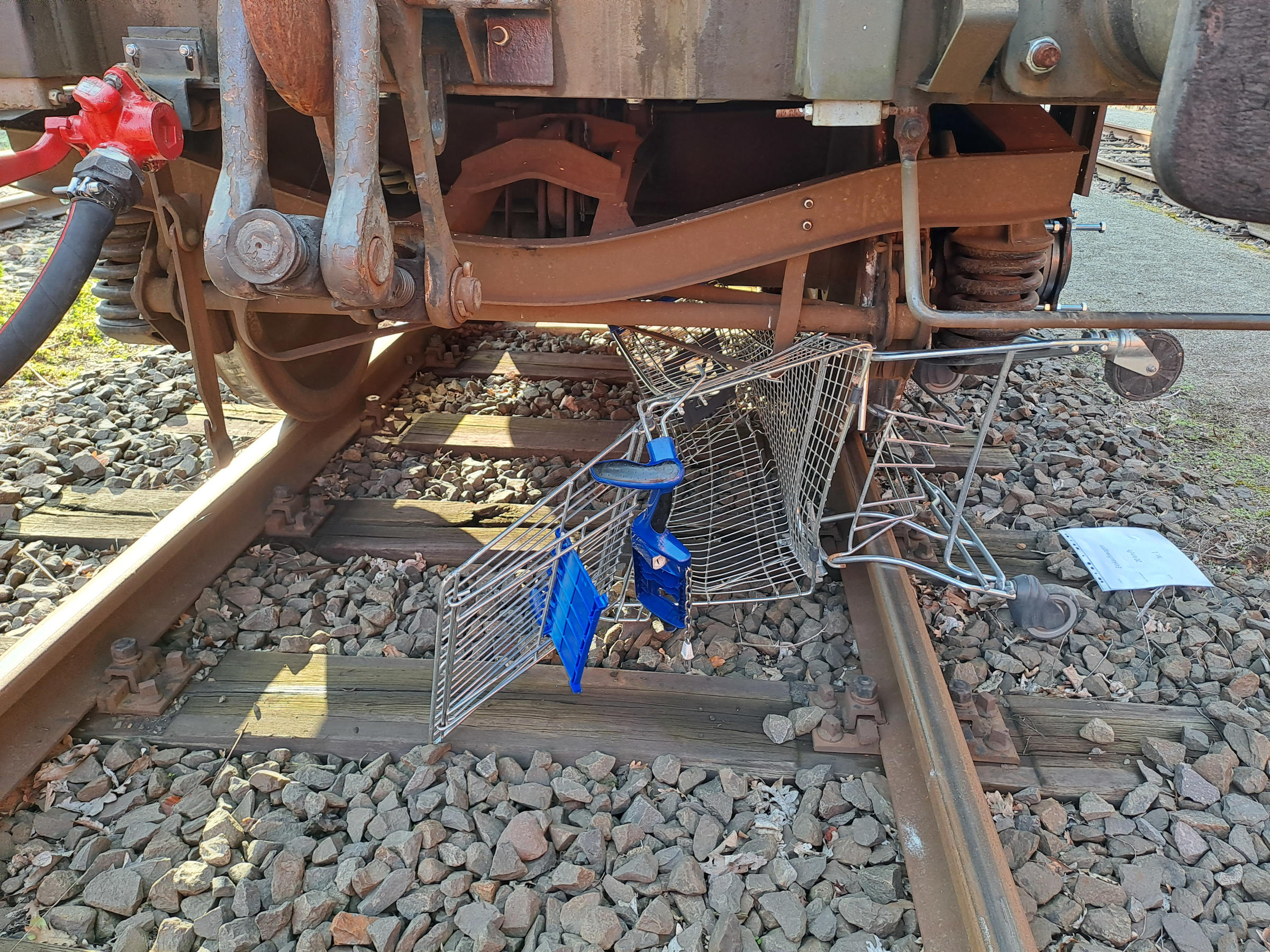}
    \caption{Driving-over experiments.}
    \label{fig:DrivingExperiments}
\end{figure}

Driving-over tests were carried out to capture real driving-over events.
The tests were conducted with two vehicles to generate events for training detection algorithms. The first vehicle was a Res freight wagon \cite{2025-Driving-Over} (cf. Figure \ref{fig:DrivingExperiments}) the second vehicle was the aTL. The Res freight wagon enabled experiments that were not allowed to conduct with the aTL due to safety regulations. For selected object classes, identical tests were carried out with both vehicles, using the same object and speed. This allows the direct comparison and filtering out of vehicle-specific effects. Both vehicles had the same sensor configuration. Within eight test days, 197 driving-over events (aTL and Res wagon) were performed on the private test track of Havelländische Eisenbahn AG (HVLE) in Berlin-Spandau. The following object classes were tested with the Res freight wagon, selected according to the probability of real world occurrence: bones, biofidel chicken, birch wood, roof batten, roof tile, overhead line, flat steel, steel wedge, steel pipe, concrete block, shopping cart (cf. Figure \ref{fig:DrivingExperiments}), bicycle, and braking shoe.

\subsection{Laboratory Experiments} \label{Laboratory Experiments}
Laboratory tests were conducted on the underrun guard and front flaps. The goal was to gather initial sensor data, validate the simulation models and provide a training dataset for first ML-based impact detection.
\subsubsection{Striking Pendulum Experiment}
\begin{figure}
    \centering
    \includegraphics[width=0.8\linewidth]{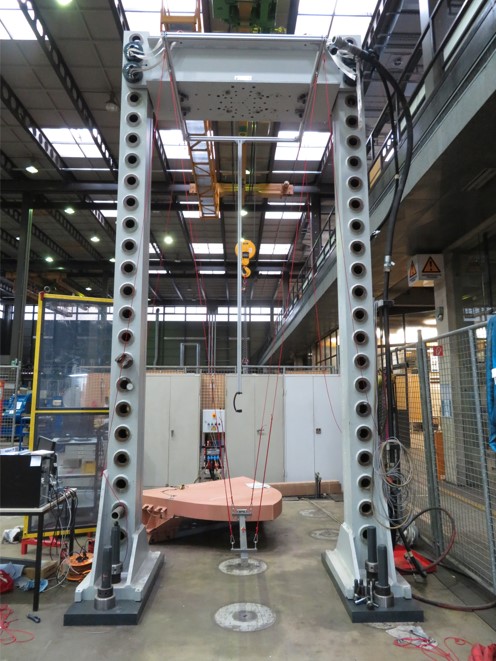}
    \caption{Structure of the stricing pendulum experiment.}
    \label{fig:ImpactPendulum}
\end{figure}

The striking pendulum experiments (cf. Figure \ref{fig:ImpactPendulum}) were designed to generate realistic acceleration signals. By reproducing controlled collision events, the experiments created reference data against which simulated acceleration signals can be compared.

The experimental setup consists of a pendulum mounted on a test portal which was released from defined heights to strike the underrun guard, which is fixed to the ground in an inverted position for easier mounting. This can be seen in figure \ref{fig:ImpactPendulum}. Two types of the impactors were deployed: a 30.5 kg steel pendulum representing stiff collision partners such as tools, rocks, or containers, and a 28.5 kg gelatin block simulating the mechanical properties of biological tissue, which is especially relevant for railway applications involving collisions with animals. The impact velocity was controlled through the pendulum’s release height, enabling repeatable tests at speeds up to approximately 4,2 m/s. Acceleration signals were recorded using piezoelectric sensors of different ranges (2,000 g and 10,000 g), ensuring coverage of both low- and high-magnitude events.

The results demonstrate that even under nominally identical conditions, acceleration signals can vary significantly, most likely due to sensor sampling rate limitations in capturing very high-frequency vibrations. Comparisons between the different sensor types revealed variations in signal resolution and level of detail. The higher-resolution sensor recorded more pronounced local extrema. Moreover, the choice of the impactor produced distinct response patterns: steel pendulum impacts caused sharp, high-magnitude accelerations exceeding 500 g, while gelatin impacts produced much lower signals, typically below 1.5 g. These differences directly reflect the stiffness contrast between metallic and biological-like materials.

Overall, the experiments confirm that the striking pendulum setup reliably generates realistic and diverse acceleration data under controlled conditions. The resulting datasets were well-suited both for FE model validation and for training AI algorithms to distinguish between various collision scenarios.

\subsubsection{Deformation Experiment}
Quasi-static deformation experiments were conducted to validate the FE models of the underrun guard by generating real DMS signals under controlled loading conditions. Two complementary objectives are pursued: (1) Testing the linear-elastic range using nondestructive quasistatic loading; (2) Test the failure behavior of the structure in the nonlinear-plastic range through forced destruction.

\begin{figure}
    \centering
    \includegraphics[width=0.8\linewidth]{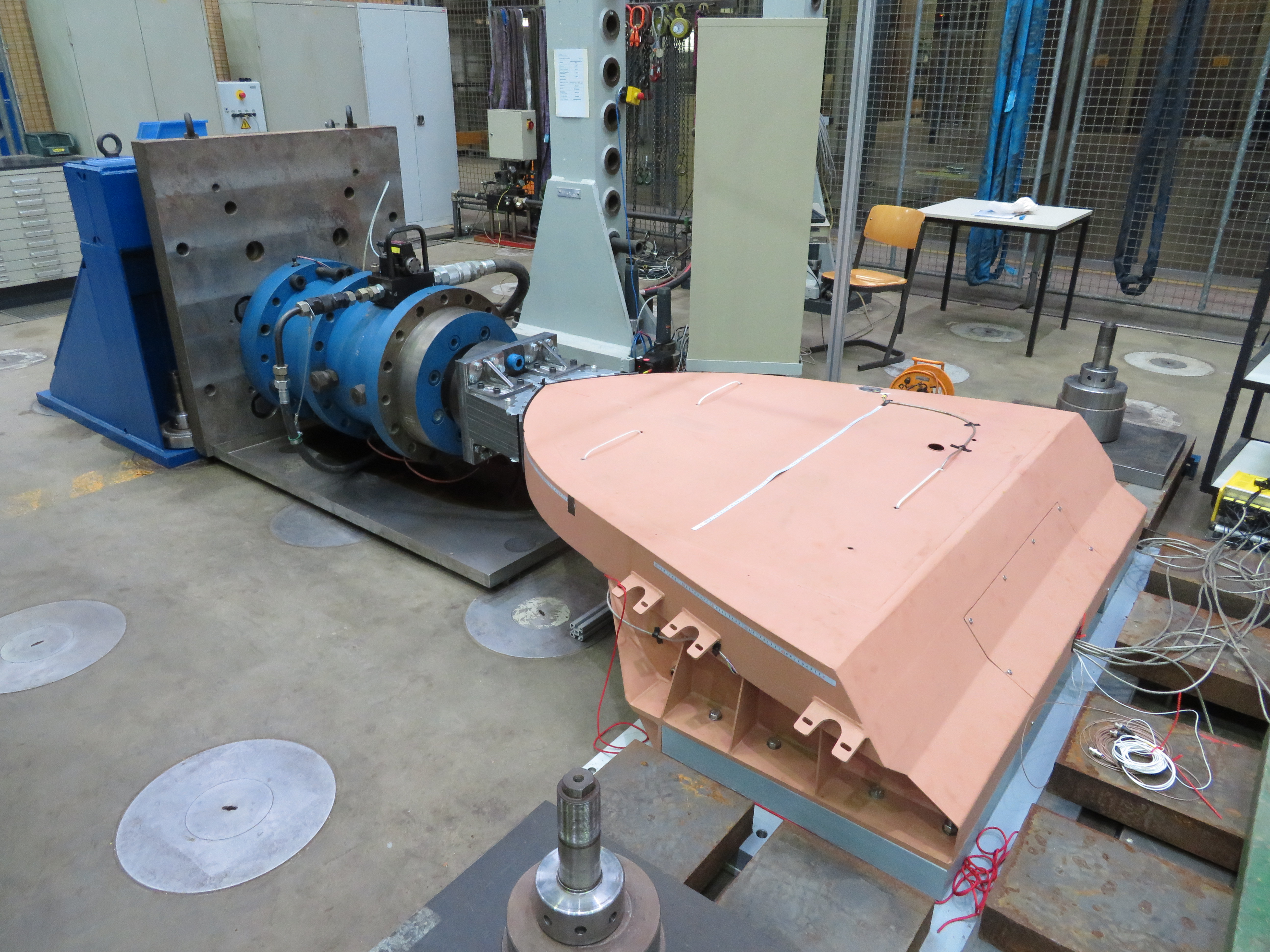}
    \caption{Setup of the quasi-static experiment.}
    \label{fig:QuasiStatic}
\end{figure}

The experiments were carried out on a test stand where the underrun guard is mounted identically to the setup of the striking pendulum experiment. The load is applied by a hydraulic cylinder through a specially designed force introduction element in accordance with DIN EN 15227 \cite{DINEN15227_2011}. This setup is shown in figure \ref{fig:QuasiStatic}.

For the quasistatic tests, the cylinder is force-controlled and loaded to 60 kN, which is then held over the test duration to record elastic strain responses without damaging the structure.
For the nonlinear-dynamic tests, the cylinder is displacement-controlled with a predefined stroke of 92 mm, corresponding to the maximum capacity of the hydraulic actuator. In this case, the test is performed at variable speed, and the load increases until either the cylinder reaches its force limit or the structure fails.

\begin{figure}
    \centering
    \includegraphics[width=0.8\linewidth]{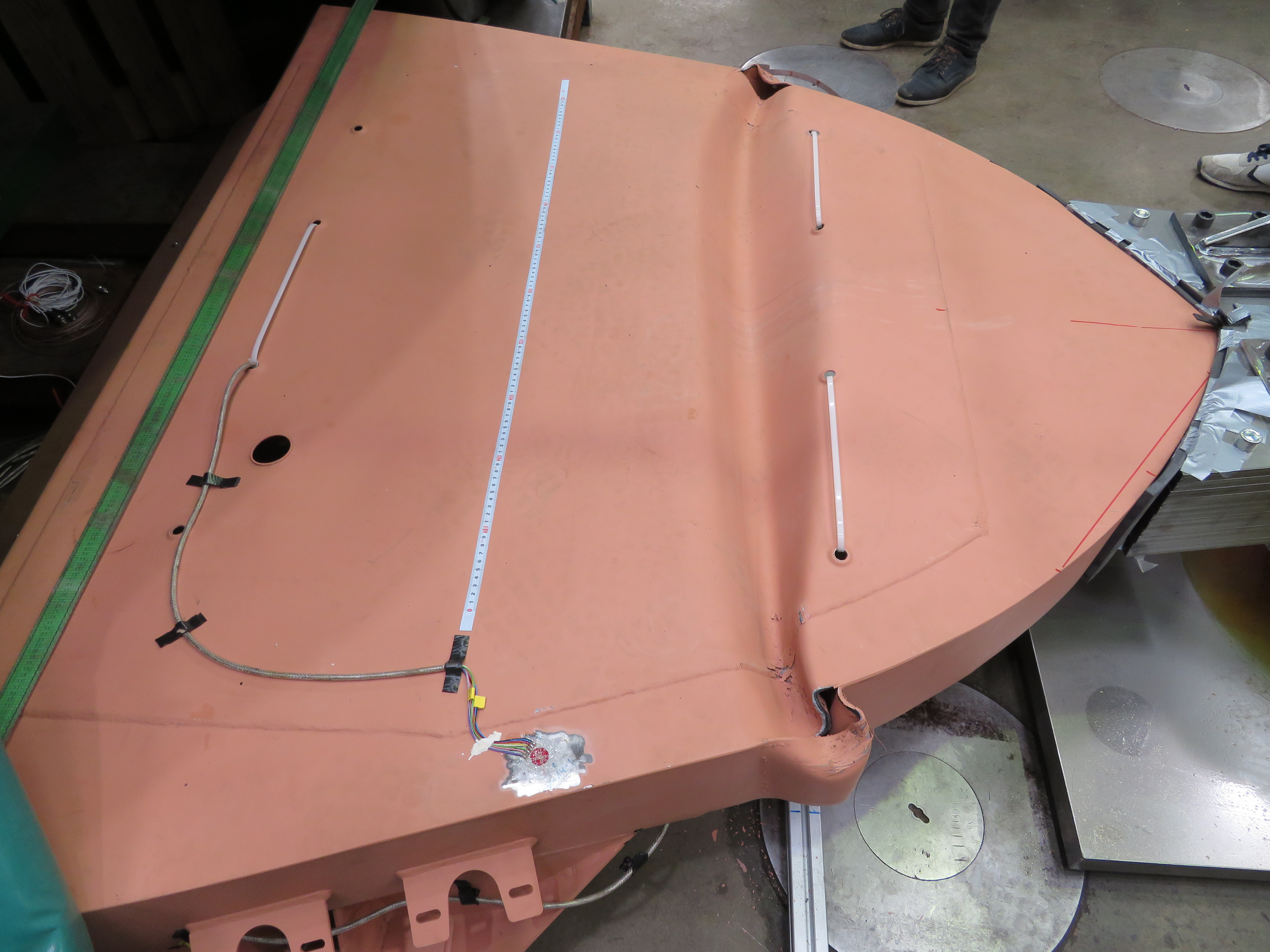}
    \caption{Damage Assessment after the destructive quasi-static non-linear test.}
    \label{fig:DamageQuasi}
\end{figure}

In the quasistatic tests, clean and reproducible DMS signals were obtained in the elastic regime, providing a reliable references.
In contrast, the nonlinear-dynamic tests required multiple repetitions before full structural failure could be achieved. 
During the second run, the cylinder reached a peak force of 889.1 kN, leading to significant plastic deformation but not complete destruction of the underrun guard. In the subsequent repetition of the fourth run, the structure failed at a slightly lower force level of 875.9 kN due to pre-existing damage. The failure mode involved pronounced plastic buckling, cracking of weld seams, and local tearing in the vicinity of the DMS positions. This can be seen in figure \ref{fig:DamageQuasi}.

Overall, the quasi-static tests confirm the suitability of the experimental approach for generating high-quality reference data for both elastic and plastic deformation ranges. The DMS signals recorded under defined load conditions provide essential input for FE model validation, while the observed failure mechanisms supply a direct comparison to simulation-predicted damage patterns. This ensures that the models not only reproduce strain distributions accurately, but also correctly capture the onset and progression of structural failure, thereby strengthening the basis for reliable damage assessment.

\subsubsection{Front Flap Experiment}
Three types of impact tests were performed on the front flaps, (1) 3-point-bending test (2) drop test and (3) projectile impact test, where object were shoot at the front flap. 
These tests were an important part of obtaining learning data sets. Preliminary tests, i.e. the quasi-static 3-point bending test and the drop tests from a height of 3 m, were carried out first. These provide insights into material behavior and initial learning data. The 3-point bending tests confirmed the quasi-isotropic laminate structure and enabled the determination of important material parameters. Drop tests showed the influence of stiffness and sensor position on acceleration signals, and validated the dynamic behavior of the nose flap and generated initial training data.

\begin{figure}
    \centering
    \includegraphics[width=1\linewidth]{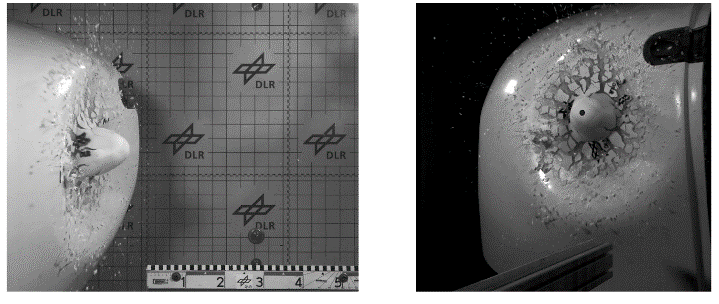}
    \caption{Impact of a 1.8 kg dummy bird at 100 m/s.}
    \label{fig:Impact}
\end{figure}

At the projectile impact test, gravel stones (50 g and 200 g), ice balls (hail) (64 g) and ellipsoidal artificial birds DLRRAB Mk 2.3 (cf. Fig. \ref{fig:Impact}) weighing 70 g, 454 g and 1800 g were used. These were projected onto the nose flap at speeds of 30 m/s, 60 m/s and 100 m/s. Depending on the mass and composition of the projectile, the nose flap deforms to a corresponding degree. This deformation causes a wave in the material that propagates through the component, which in turn leads to damage to the nose flap depending on the severity of the deformation \cite{WinklerHoehn2025b}. Such an impact event is shown in figure \ref{fig:Impact}.

Damage to the front flap caused by artificial bird impacts occurred mainly at the coupling opening and at the connection brackets to the mimic. Only minor damage was observed in the immediate impact area. Gravel stone impacts showed a different behavior, with the primary damage located directly at the point of impact. Ice ball impacts did not produce any visible damage.

\begin{figure}
    \centering
    \includegraphics[width=1\linewidth]{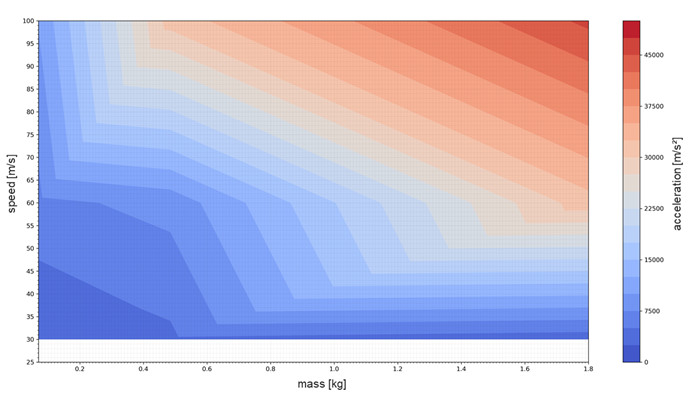}
    \caption{Acceleration-speed-mass area diagram. Blue tones represent low accelerations (below 22,5 m/s²) and red tones represent higher accelerations up to 50 m/s².}
    \label{fig:SpeedMass}
\end{figure}

Based on the acceleration signal, it is not possible to identify the type of projectile. However, an impact can be detected regardless of the position of the sensor. In order to identify the type of object that has struck or to classify it, additional sensors are needed. By knowing the peak of the acceleration signal, the speed of the aTL and the type of object, the mass of the object and the corresponding damage can be determined as shown in figure \ref{fig:SpeedMass}. The impact tests provide valuable data sets for the training of machine learning models \cite{WinklerHoehn2025b}.

\subsection{Simulations} \label{Simulations}
This chapter will show the results of the simulation for the underrun guard and for the front flaps. The conducted simulations demonstrate that the applied methodology is effective in generating realistic sensor data and predicting the structural response. The results of the simulation are verified and validated by the experimental tests and were used to generate collision and impact events for creating a damage library and training ML algorithms.

\subsubsection{Simulation Underrun Guard}
The conducted simulations demonstrate the structural response of the metallic load-bearing structure under collision events. The simulated datasets included DMS signals and acceleration measurements, capturing both time-dependent strain and dynamic acceleration responses. DMS proved particularly useful for monitoring the structural condition, as they reflect deformations within the material, while acceleration sensors provided complementary information for impacts involving heavy or rigid objects. However, it was observed that in regions where welds are likely to fail during a collision, the simulated strain gauge signals do not accurately represent the physical response, indicating a limitation of purely model-based data generation.

\begin{figure}
    \centering
    \includegraphics[width=0.9\linewidth]{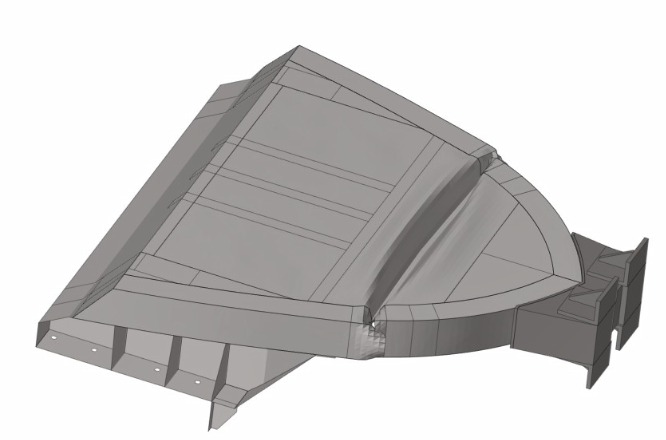}
    \caption{Damage assessment of the validation simulation.}
    \label{fig:SimulationDamage}
\end{figure}

The structural analysis of the simulated collision events revealed a clear correlation between the mass and type of the impacting object, the collision velocity, and the resulting damage severity. Minor collisions, such as impacts with lighter objects, predominantly led to small plastic deformations, dents, or minor bending. In contrast, heavier or more rigid collision partners caused severe damage, including significant plastic deformations, fractures, and cracks, particularly at welds and critical structural points. Undamaged cases were rare or absent for the analyzed scenarios, highlighting the vulnerability of certain structural elements under high-impact loads. Figure \ref{fig:SimulationDamage} shows the validation simulation with a similar damage behavior as in the laboratory experiments (cf. Figure \ref{fig:DamageQuasi}).

In total, the simulation successfully achieves excellent qualitative agreement, accurately reproducing the location, shape, and type of damage, including a highly precise prediction of the maximum acceleration, which deviates by only $1.8\%$. While the quantitative accuracy for detailed strain measurements shows room for improvement with deviations up to $34.9\%$, the model provides a fundamentally good and reliable representation of the resulting structural state.

The results underscore the importance of considering potential structural failures, such as weld fractures, when interpreting sensor signals, and demonstrate the value of combining different sensor types to achieve a complete understanding of structural behavior under collision loads.

\subsubsection{Simulation Front Flaps}
\begin{figure}
    \centering
    \includegraphics[width=1\linewidth]{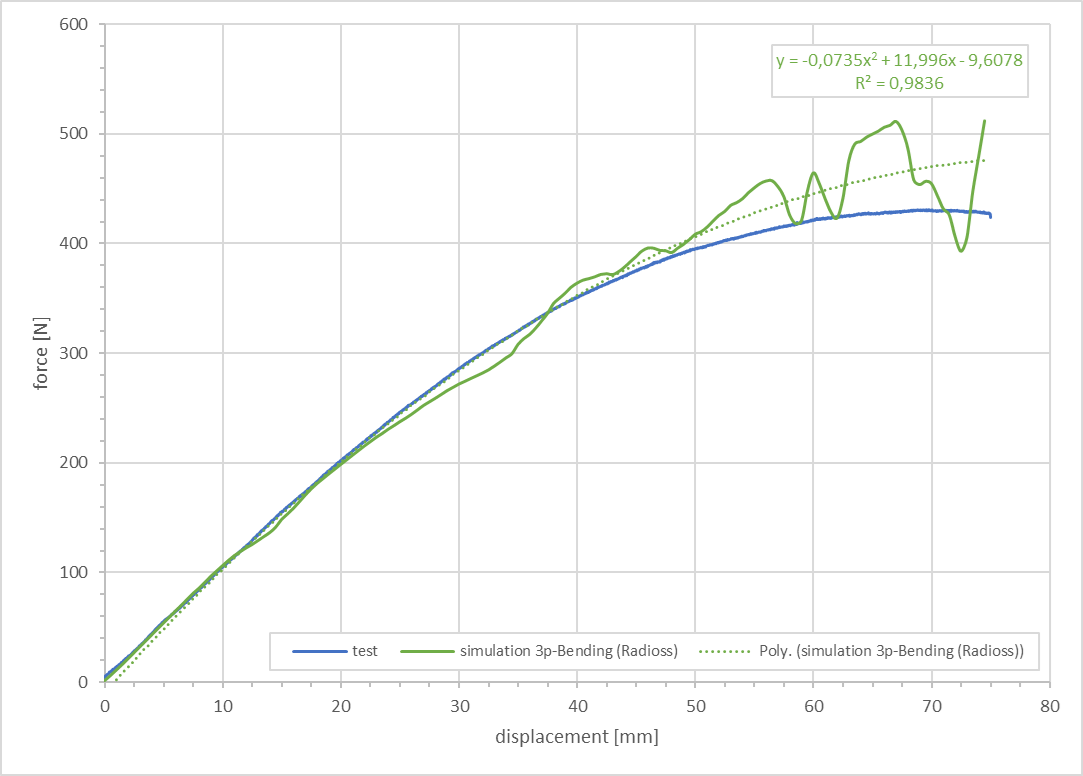}
    \caption{Comparison of the force-displacement curves from the 3-point bending tests with corresponding FE simulations.}
    \label{fig:ForceDisplacement}
\end{figure}

The numerical tests carried out with clearly defined loads are intended to serve as simulation-based verification and validation of the experimental tests and thus the sensor network. These were also used for data evaluation using AI methods. The simulation of the 3-point bending tests showed good correlations between the simulations and the tests. The deviations from the tests are between -1.8 \% and 0.64 \% for small deformations as seen in figure \ref{fig:ForceDisplacement}.

\begin{figure}
    \centering
    \includegraphics[width=1\linewidth]{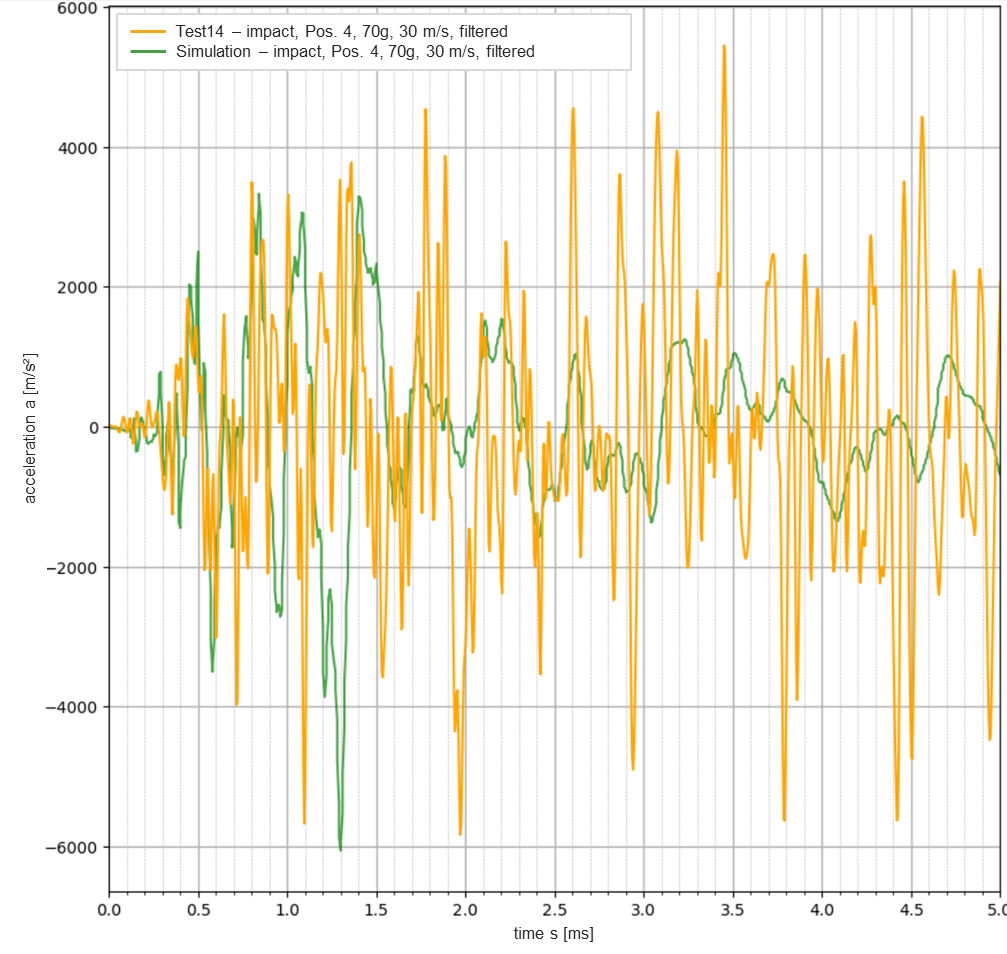}
    \caption{Comparison of acceleration signals from simulation and tests at sensor position 4.}
    \label{fig:ComparisonAcceleration}
\end{figure}

The deviations increase for larger deformations. This can be attributed to the elastic material model used for simplifications. The numerical tests confirm the selection of material parameters for the material used in the nose flap \cite{WinklerHoehn2025b}. 
In the following, the material parameters validated by the simulations are used for the simulations of the impact tests. These serve to verify the experimental impact tests.
The simulations captured the dynamic response of the impact body accurately enough, with only minor differences in the first peak amplitude. In addition, a similar damping profile is available \cite{WinklerHoehn2025b}.
The signals from the impact tests are significantly noisier than those from the drop tests. This makes it difficult to evaluate the exact amplitude difference between simulations and tests. It can be emphasized that the amplitudes are similar up to approximately 1.5 ms, as shown in figure \ref{fig:ComparisonAcceleration}. Deviations can be attributed to the modeling parameters used for the nose flap. Previous investigations have shown that the impact occurs and is completed within the first 0.5 ms. Thus, the simulation can be considered sufficiently validated for the range from 0.5 ms to 1 ms inclusive. This confirmed the reliability of the model for capturing the initial impact behavior \cite{WinklerHoehn2025b}.
In summary, the simulations of the 3-point bending, drop and impact tests showed good consistency with the experimental results.

In summary, the simulations provide a validated basis for further analysis and prediction of the mechanical behavior of the nose flap under various load and impact conditions \cite{WinklerHoehn2025b}.

\section{Developed Methods and Results}
In the following the results for the main use cases are outlined, i.e. (1) impact and driving-over detection, (2) condition-based maintenance, and (3) optimized vehicle design.

\subsection{Impact Detection}

\begin{figure}
    \centering
    \includegraphics[width=1\linewidth]{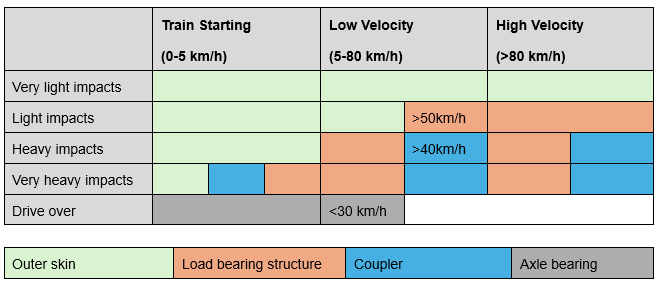}
    \caption{Detection capabilities based on simulations and experiments. \textbf{Very light impacts}: birds, branches/small tree; \textbf{Light impacts}: human, small animal
    \textbf{Heavy impacts}: big animals, big tree, car
     \textbf{Very heavy impacts}: truck, train
    \textbf{Driving over}: rigid objects 20mm.
}
    \label{fig:ImpactResults}
\end{figure}

The detection of impacts at the front of the vehicle can be realized on the basis of signals from the acceleration sensors. 
To determine the detection accuracy, the impact events from the experiments (cf. Section \ref{Laboratory Experiments}), and simulations (cf. Section \ref{Simulations}) were compared with the signals from regular operation. To improve signal quality and therefore detectability, all sensor signals, including the signals from the simulations, were processed with a CFC1000 filter. The usage of this filter provided a better comparability between the real and the simulated signals. The filtered signals were then used to distinguish normal operation signals from impact events on the front and side of the train (cf. Figure \ref{fig:ImpactResults}).

The event classification is based on a thresholding approach. For each event, defined as a time series within a specified interval, the maximum value is compared against a threshold. Values exceeding or equaling the threshold are classified as positive; otherwise, the event is classified as negative. The classifier is trained by determining an appropriate threshold using laboratory recordings and simulation results for positive events and regular aTL operation data for negative events. Threshold optimization is performed via a global optimization method \cite{BlackBoxOptim}, employing a fitness function that minimizes misclassifications errors (0 for correct classification, +1 for false classifications) with an increased penalty (+10) for false positives. False positive detections shall be avoided as they would lead to far-reaching disruption of operations in real-world scenarios. For each acceleration sensor, five classifiers were defined, corresponding to peak-to-peak acceleration, RMS acceleration, frequency band acceleration, velocity peak-to-peak, and velocity RMS. Finally, an ensemble classifier aggregates the outputs of the five classifiers using majority voting to improve the robustness of impact detection.

The method achieved a detection rate of 90.09\% for real impact events from the laboratory and 100\% for simulated impact events, with no false positives.
 The detection capabilities are shown in figure \ref{fig:ImpactResults}. The outer skin is able to detect light to heavy impacts depending on the speed. At higher velocities the load bearing structures and the coupler can detect impacts while the outer skin is completely damaged. 

\subsection{Driving-over Detection}

For detecting driving-over events, the acceleration sensors at the axle bearing were used.
The detectability of a wheel driving over an object  particularly high if the wheel is lifted sufficiently when rolling over the object so that it then falls back onto the rail. The highest amplitudes are not caused by the immediate impact with the object, but by the wheel reestablishing contact with the rail surface. The distinction between such events and regular track-related vibrations depends largely on the quality of the rail infrastructure. On heavily frequented main lines with regularly maintained, smooth running surfaces, significantly lower vertical accelerations occur than on less frequented branch lines or in shunting areas. 
Two detection algorithms have been developed. One classical and one deep learning based approach. It could be shown that in a shunting yard a mean detection accuracies for the classical approach of 88,6\% and for the deep learning approach of 99,6\% could be reached \cite{2025-Driving-Over}. 
However, this method reaches its limits on tracks with low maintenance, high wear, or uneven running surfaces. In such cases, only very large events can be detected. 

\subsection{Generated Damage Library and Damage Assessment}

The simulations were conducted to generate training data for an algorithm designed to identify collisions and damage to the vehicle structure in real-world rail operations and recommend appropriate actions. This required realistic collision scenarios with realistic collision partners that could occur or are likely to occur in actual rail operations. Documented collision events were first researched, i,e, accident reports, media reports. Furthermore, interviews with maintenance experts and train drivers were conducted.

Based on documented collision events, surrogate models of typical collision partners with a maximum mass of 1.5 t were selected to reduce FEM computational effort. The simulated partners comprised (1) living beings, represented by a wild boar (150 kg cylinder, 0.37 m × 1.3 m) to reflect frequent wildlife impacts \cite{Jumin2017Protection, Broms2002Finite, Gens2001Moose}; (2) trees, modeled as coniferous (665 kg, 0.3 m × 12 m) and deciduous trunks (914 kg, 0.3 m × 12 m); and (3) heavy, hard objects in accordance with DIN EN 15227 \cite{DINEN15227_2011}, represented by generic steel geometries—0.05 t cubes (0.3 m side), 0.25 t cylinders (0.9 m × 0.6 m), and 1.5 t cuboids (4 m × 1.8 m × 0.55 m).

Fifteen load cases were simulated, categorized into “Living beings”, “Trees”, and “Heavy, hard objects”. Parameters such as impact speed (40–160 km/h), orientation (transverse or 30° offset), and impact position (centered or offset) were systematically varied. The resulting dataset, comprising sensor responses and structural states, covers a representative spectrum of relevant collision scenarios.

The simulation and experimental results were systematically organized in a database according to the magnitude of the acting forces. The severity of the structural damage generally increased proportionally with the impact force. However, this relationship holds true only when each structural component is considered individually. Subsequently, domain experts assessed the resulting damage patterns and determined suitable operational and maintenance measures for each case. This approach enables the derivation of correlations between impact forces, damage severity, and corresponding response measures, providing a foundation for future predictive decision-making in real-world operations.

\subsection{Optimized Vehicle Design and Load Spectra} 

\begin{figure}
    \centering
   \includegraphics[width=1\linewidth]{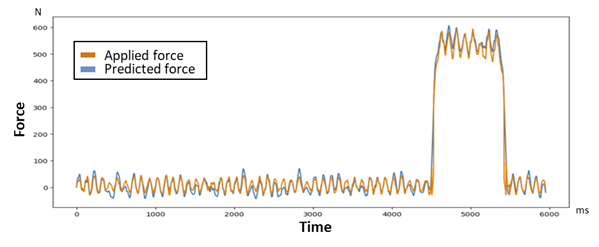}
    \caption{Example of an applied force on the FE model and the predicted force by the trained neural network.}
    \label{fig:force-time}
\end{figure}

Outdated standards can often lead to oversized components and thus to an increase in mass. In order to meet our requirements for lightweight construction, the components can be designed according to the loads that occur during operation. This has the advantage that the components can be designed according to the real load, thus leading to an overall reduction in weight. In our case, the installed sensor system is used to determine the loads applied. The main objective is to use these for FE simulations or fatigue analyses.
Forces are the easiest loads to define in an FE model. 

However, due to nonlinear effects, these loads cannot be reliably derived from the installed sensor data. To address this limitation, an AI-based method was developed to solve the inverse problem.

The sensor data are provided as input to a neural network, which estimates the corresponding force. For training the network, a method was developed to automatically generate a large dataset of acceleration–force and strain–force pairs from FE simulations (patent: 10 2024 100 677.3) \cite{Laporte2024}. In this process, forces were applied to an FE model, and the resulting structural responses, namely accelerations and strains, were recorded at the sensor locations. Figure~\ref{fig:force-time} presents an example of an applied force in the FE model and the force predicted by the trained neural network.

\begin{figure}[t] 
    \centering
   \includegraphics[width=1\linewidth]{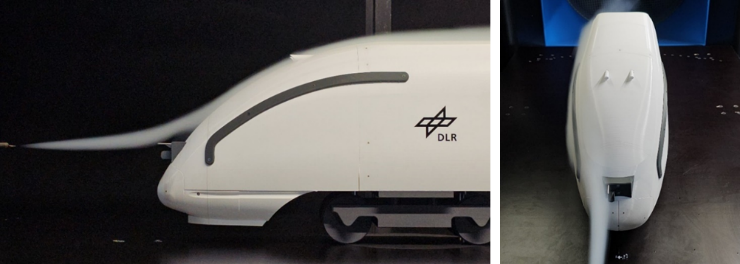}
    \caption{Wind tunnel experiments with a small scale train.}
    \label{fig:windtunnel}
\end{figure}

To include the aerodynamic forces affecting the train, small-scale wind tunnel tests at 1:10 and 1:25 scales were conducted to calibrate the nose-mounted dynamic pressure probe and determine the drag, lateral force, and lift for the first car under various geometric configurations (cf. Figure \ref{fig:windtunnel}) \cite{Buhr2024}. These corrected aerodynamic measurements, combined with surface pressure data and an operational database of strain gauge and acceleration sensor measurements, enabled a load spectrum analysis for real-world operating conditions.

\section{Conclusion and Future Work}
The KI-MeZIS project demonstrated that integrating advanced sensor systems enables AI-based analysis for impact and driving-over detection, condition-based maintenance, and optimized rail vehicle design. By combining field tests, laboratory experiments, and simulations, the project established a comprehensive data foundation to support predictive maintenance, improve operational safety, and guide vehicle design optimization. Additionally, the project provides a technological basis for future rail automation and life-cycle management.

Building on these results, future research will focus on automating the simulation of numerous underrun guard crash scenarios and developing machine learning models to quantify structural damage and translate it into operational guidance. The goal is to link technical damage assessment directly to maintenance decisions, advancing AI-driven decision support in railway operations.

\section*{Acknowledgment}

The authors would like to thank Michael Rohrschneider and Martin Zeitz (DB) for their support during the test runs, as well as for their contributions to system integration and approval. The supervision of the system installation in the aTL by Mario Scherbaum (MSG) is gratefully acknowledged. Furthermore, the authors extend their thanks to Jörg Jacobs, Kai Sonnenburg, and Florian Hoyer (DB Systemtechnik) for their work on the design and integration support. The valuable contributions of Carl-Jonas Braun (IMA), Arne Henning, James Bell, Dorothea Schlie (DLR), Matthias Härter (DB InfraGO), Florian Stark, Nikolay Chenkov, and Rubens Rossi (IA GmbH) throughout the project are also sincerely acknowledged.

\IEEEtriggeratref{1}
\IEEEtriggercmd{\vspace{+\baselineskip}}
\bibliographystyle{IEEEtran}
\bibliography{IEEEexample}

\end{document}